\documentclass{article}

\usepackage{arxiv}
\usepackage[utf8]{inputenc} 
\usepackage[T1]{fontenc}    
\usepackage{hyperref}       
\usepackage{url}            
\usepackage{booktabs}       
\usepackage{amsfonts}       
\usepackage{nicefrac}       
\usepackage{microtype}      
\usepackage{lipsum}		
\usepackage{doi}
\usepackage{graphicx}
\usepackage{epstopdf}
\usepackage{subcaption}
\usepackage{framed,multirow}
\usepackage[english]{babel}

\title{Self-Supervised Face Presentation Attack Detection with Dynamic Grayscale Snippets}

\date{} 					

\author{ \href{https://orcid.org/0000-0001-7191-0245}{\includegraphics[scale=0.06]{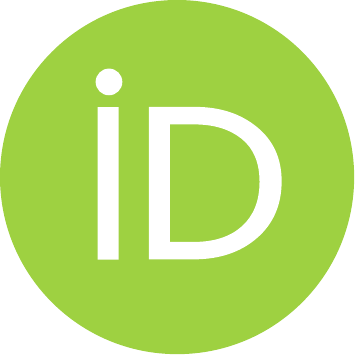}\hspace{1mm}Usman Muhammad and Mourad Oussalah}\thanks{Corresponding author: Usman Muhammad} \\
	Center for Machine Vision and Signal Analysis\\
	University of Oulu\\
	\texttt{Muhammad.usman@oulu.fi} \\
}

\renewcommand{\shorttitle}{\textit{arXiv} Template}

\hypersetup{
pdftitle={Self-Supervised Face Presentation Attack Detection with Dynamic Grayscale Snippets},
pdfsubject={q-bio.NC, q-bio.QM},
pdfauthor={Usman Muhammad, Mourad Oussalah},
pdfkeywords={First keyword, Second keyword, More},
}

\begin{document}
\maketitle

\begin{abstract}	
Face presentation attack detection (PAD) plays an important role in defending face recognition systems against presentation attacks. The success of PAD largely relies on supervised learning that requires a huge number of labeled data, which is especially challenging for videos and often requires expert knowledge. To avoid the costly collection of labeled data, this paper presents a novel method for self-supervised video representation learning via motion prediction. To achieve this, we exploit the temporal consistency based on three RGB frames which are acquired at three different times in the video sequence. The obtained frames are then transformed into grayscale images where each image is specified to three different channels such as R(red), G(green), and B(blue) to form a dynamic grayscale snippet (DGS). Motivated by this, the labels are automatically generated to increase the temporal diversity based on DGS by using the different temporal lengths of the videos, which prove to be very helpful for the downstream task. Benefiting from the self-supervised nature of our method, we report the results that outperform existing methods on four public benchmarks, namely, Replay-Attack, MSU-MFSD, CASIA-FASD, and OULU-NPU. Explainability analysis has been carried out through LIME and Grad-CAM techniques to visualize the most important features used in the DGS.
\end{abstract}


\section{INTRODUCTION}
Facial recognition technology has been employed for numerous real-world applications, such as airport passenger screening, mobile phones, banking, and law enforcement surveillance. However, recent studies show that face recognition systems are vulnerable to presentation attacks, such as silicone masks, replay of video, and photo attacks. In particular, these spoofing attacks pose a great threat to face recognition systems and can throw off the technology. To protect the face recognition systems against spoofing attacks, many software-based or hardware-based methods have been introduced \cite{li2022face, zhang2020casia, wang2022domain, liu2021face}. These methods provide an improved detection but their performances degrade drastically under real-world variations (e.g., the quality of video recording devices and illuminations). Thus, facial recognition technology needs to be reinforced with presentation attack detection (PAD) techniques to secure the existing face biometric systems.
\begin{figure}
\centerline{\includegraphics[width=9.50cm]{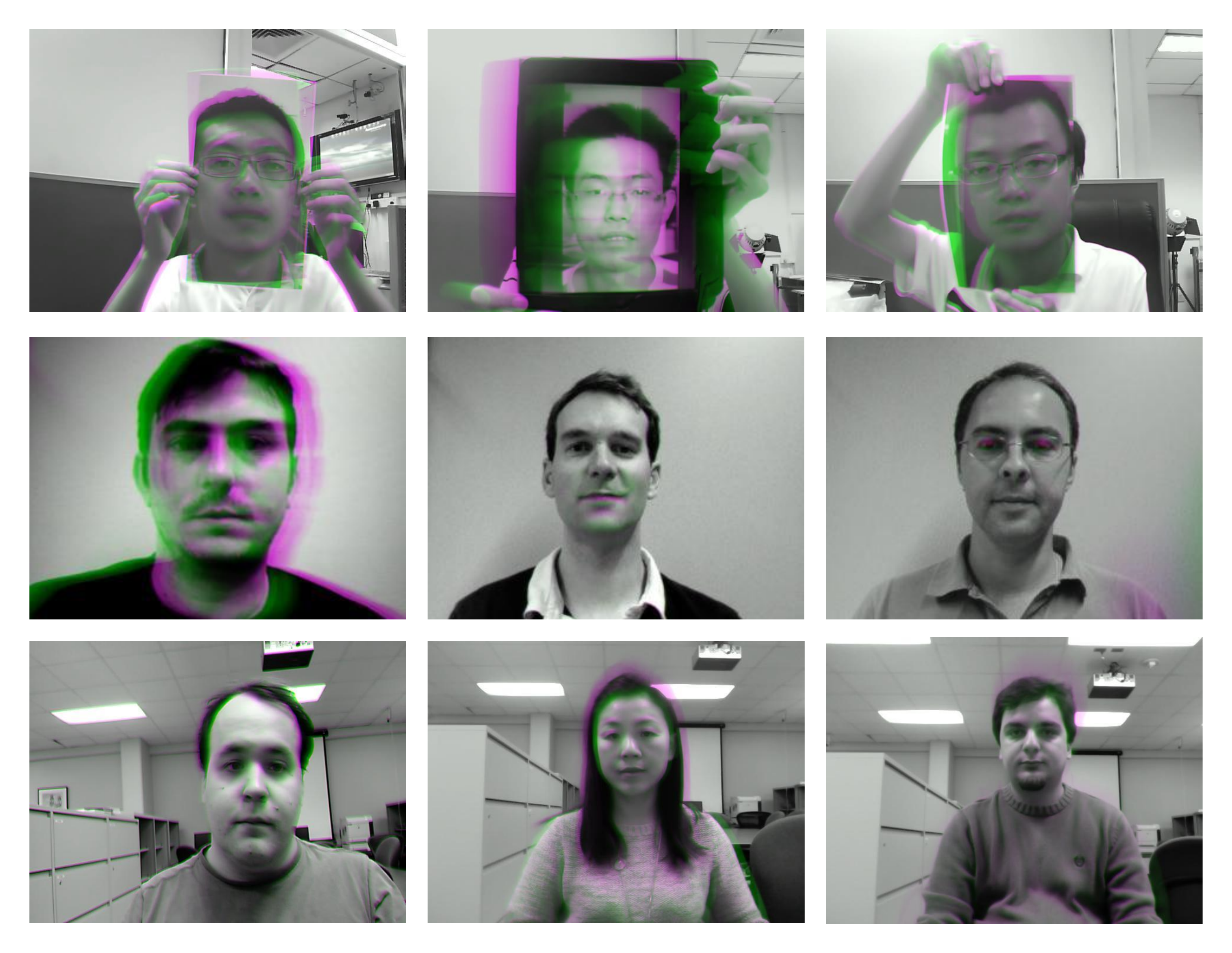}}
\caption{Visualization of example images (DGS) from CASIA, Replay-Attack, and MSU datasets. Colored areas show the corresponding pixel displacement along the time-axis.}
\end{figure}
One way to address the problem of spoofing consists of using temporal feature learning. The existing countermeasures can be roughly divided into three main categories: (i) extracting dynamic features through CNN network, \textit{e.g.,} based on optical flow \cite{li2022face}, (ii) using Spatio-temporal features based on two-stream CNN \cite{kim2020action}, and (iii) learning long-range (sequential) data \textit{e.g.,} through Recurrent Neural Networks (RNN) \cite{muhammad2022self}. Indeed, the widespread adoption of temporal feature learning for face PAD in supervised settings has achieved excellent results. However, the success of supervised feature learning relies on massive amounts of manually labeled data, which is expensive and requires great effort. Alternatively, a prominent paradigm is the so-called self-supervised learning (SSL) which aims to empower machines without explicit annotations and has shown potential in both image and video domains \cite{schiappa2022self}. SSL can be separated into two main categories: exploiting context and videos. 

Various methods have been proposed in context-based self-supervised learning by designing image-specific pretext tasks such as colorizing gray images, image in-painting, jigsaw puzzles of image patches, and rotations of images. Compared with static images, video-specific pretext tasks (e.g., predicting clip orders, time arrows, and paces) provide richer sources of “supervision” \cite{jaiswal2020survey}. Therefore, it is not trivial to expand the image-based approaches directly to the video domain due to dynamic and fine-grained movements or actions \cite{schiappa2022self}. 
\begin{figure*}
\centerline{\includegraphics[width=12.50cm]{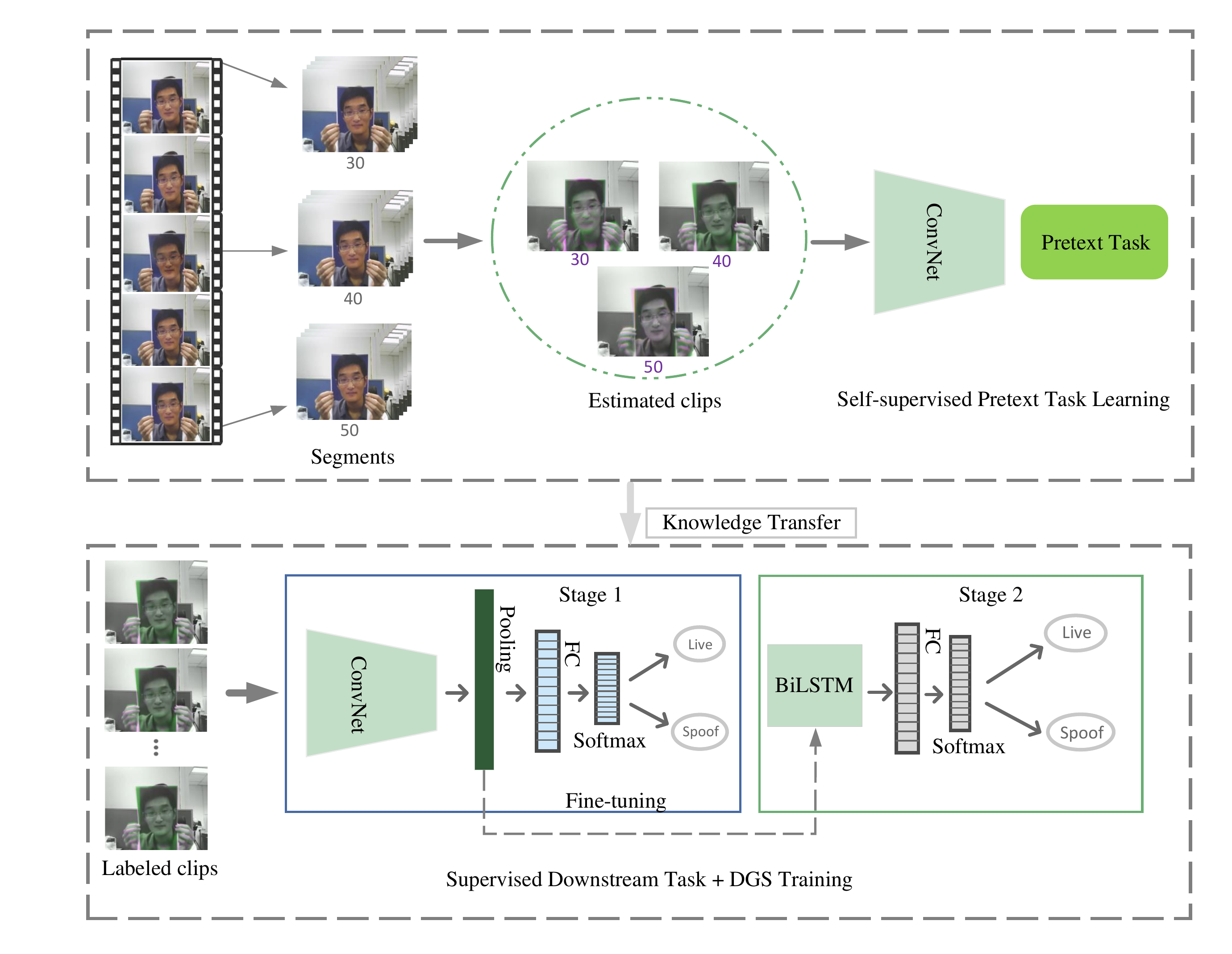}}
\caption{An illustration of the DGS-based supervised and self-supervised training tasks. A model is first trained with a pretext task with different temporal lengths to predict
the length of a given DGS encoded video segment, then fine-tuned for the actual downstream task of face PAD. For the supervised learning phase, the CNN is first fine-tuned in stage 1 and then the BiLSTM is trained using the fine-tuned features (stage 2) to provide the final PAD detection.}
\end{figure*}
In order to avoid human labeling, Walker {\textit{et al}} \cite{walker2015dense} proposed to estimate the motion information based on dense optical flow. Their approach avoids data annotation by employing a convolutional neural network (CNN) for motion prediction. However, the optical flow algorithm increases computational complexity and the memory space in dealing with video data. Usman {\textit{et al}} \cite{muhammad2022self} proposed a self-supervised learning for face PAD based on a temporal sequence sampling. In their approach, the expensive estimation of the global motion is needed, which causes to increase in the computational complexity of the method. Recently, the authors in \cite{muhammad2022adaptive} claimed that the most important patterns for determining the decision on PAD are consistent with motion cues associated with the artifacts, i.e., eye blinking, hand trembling, and head rotation. These motion cues are valuable and essential to examining real and spoofing attacks where the relative motion between the background and the face region can be vital. Although previous video-based PAD methods provide an improved performance \cite{muhammad2022self,muhammad2022adaptive}, such methods require complex feature engineering skills and increase the computational cost of the model.

Inspired by the above discussion, our goal is to develop not only a computationally efficient video representation learning method for summarizing the motion information, but also extend to define the pretext task as a motion prediction problem for self-supervised learning. Specifically, we achieve this by dividing the video into a set of non-overlapping segments. Then, for each segment, three reference frames are chosen and transformed into three single-channel (i.e., grayscale) images. The final 3-channel image which we called dynamic grayscale snippet (DGS) is formed by assigning each into three different channels such as R(red), G(green), and B(blue), separately. Fig.1 displays some example images of the DGS method. One can see that the background has grayscales due to the static background while the motion is well separable by representing colors. 

Motivated by this, we present a self-supervised video representation learning where the pretext task is to predict the lengths of the videos based on the DGS method. In particular, the proposed DGS with different temporal lengths provides supervision to predict future motion that can be used for the downstream task. It is worth mentioning that the importance of DGS is not limited to SSL and has at least three main advantages. First, DGS can be input to any 2D CNN for a still image, where “still” captures the long-term motion dynamics in the video. Second, DGS decreases the risk of over-fitting of human faces, since the PAD videos consist of hundreds of frames repeating similar face patterns. Third, the representation provides a fast approximation in comparison to features obtained by complex processing such as optical flow \cite{horn1981determining}, or global motion estimation \cite{muhammad2022self} (we will further discuss this aspect in section III). 

In summary, our main contribution is three-fold. (i) We propose a 3-channel image, termed a dynamic grayscale snippet (DGS) for motion information that enables the use of a 2D CNN for the temporal stream and reduces the computational complexity simultaneously. (ii) To mitigate the cost of collecting a large amount of labeled data, different temporal lengths based on DGS are used in self-supervised learning to complement pretext task that offers a powerful supervisory signal for video-based face PAD. (iii) The effectiveness of the proposed approach is evaluated on four standard databases and promising generalization ability is achieved especially using the official cross-test and intra-dataset evaluation protocols.

\begin{figure}
\centerline{\includegraphics[width=10.70cm]{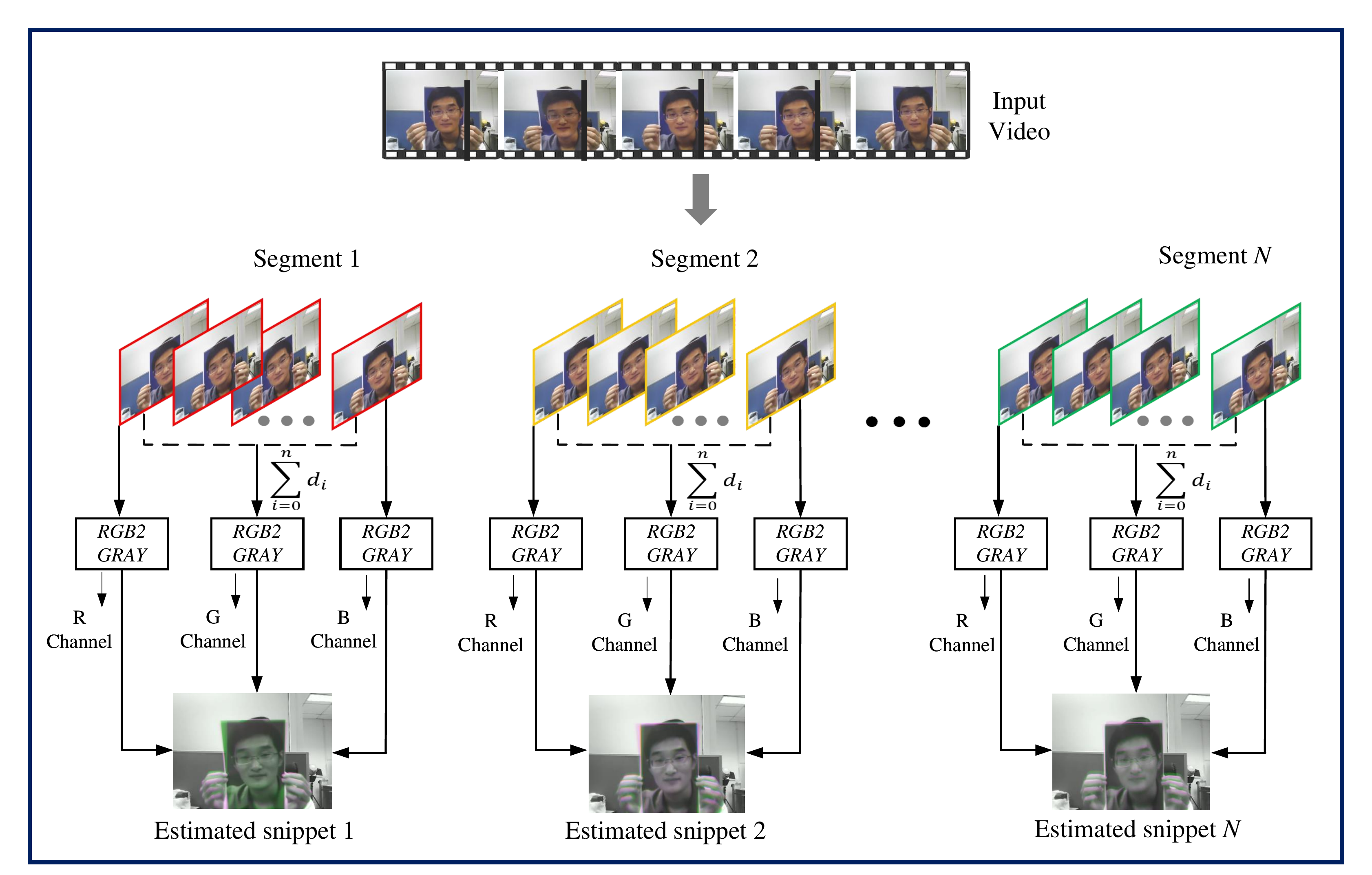}}
\caption{Formation of the proposed dynamic grayscale snippets (DGS).}
\end{figure}

\section{Proposed Method}
The backbone of the proposed self-supervised learning is the generation of DGS where the different temporal lengths are used before feature extraction. The main architecture is illustrated in Fig. 2. During the self-supervised learning phase, we provide unlabeled DGS with different temporal lengths as a pretext for task learning. In the downstream task, the learned representations are fine-tuned on labeled classes, i.e., live and spoofed faces. We first explain the procedure of estimating the dynamic grayscale snippets and then, the procedure of incorporating self-supervised learning is described.

\subsection{Dynamic grayscale snippets}
Suppose that a video $V$ is equally divided into $P$  non-overlapping segments, $i.e., V = \{S_k\}^P_{s=1}$, where $S_k$ is the $k$-th segment. The length of each segment is set to be $(X = 40)$ frames. For each segment, a dynamic grayscale snippet is formed from three selected frames and transformed into a 3-channel single image. We expect that a 2D CNN can seek the motion information from actual actions (e.g., eye blinking, hand trembling, head rotation). To achieve this, the first reference image is selected from the first frame of the segment and then converted into a grayscale image. The second image is selected based on combining (averaging) all the frames available in the segment and then assigned to a grayscale image. Similarly, the last reference image is the last frame of the segment. Since all these selected frames are converted into grayscale images, the final 3-channel single image is formed by assigning to three R(red), G(green), and B(blue) channels, respectively. By doing this, we can explore any brightness discrepancy of the corresponding pixels in the segment because it can tell us the existence of moving objects. Thus, the motion can be observed in the form of colored regions and the grayscale area will represent no motion in the image. For visualization purpose, one can see from Fig.3, that three different frames are selected along the time-axis in sequential order and then converted into grayscale images before assigning them to the final RGB image. Therefore, a pixel with an identical value remains the same in grayscale if the object does not move. In addition, the pixel in motion will change the RGB values and will have a color to depict a specific action of the moving object. These temporal snippets with different lengths are the ones used for self-supervised learning in the next stage.

\subsection{Self-supervised Representation Learning}
Given a sequence of $V$ video frames denoted as $\{l_{1}, l_{2}, l_{3}, \dots, l_{n}\}$, a set of different combinations can be formed by dividing a video into non-overlapping segments such as $P_{a} = \{w_{1}, w_{2}, w_{3}, \dots, w_{30}\}$, $P_{b} = \{q_{1}, q_{2}, q_{3}, \dots, q_{40}\}$, and $P_{c} = \{j_{1}, j_{2}, j_{3}, \dots, j_{50}\}$.  We use these three different DGS encoded temporal lengths $P_{a}$, $P_{b}$, and $P_{c}$ as a class labels to train a deep CNN for predicting the length of a given DGS encoded video segment. The loss is defined as:
\begin{equation}
L = {- {\sum_{i=1}^{n}}t_{i} \ log(p_{i})}, {\ \ \ For \  n \  classes,}
\end{equation}
where $t_i$ denotes the label, $p_i$ the softmax probability for the $i^{\mathrm{th}}$ class, and $n$ the total number of classes. After pretext task learning, the model is fine-tuned on the learned representations for the actual downstream task of face PAD.

\begin{table}[t]
\centering
\caption{The average execution times in fps for TSS images, optical flow and the proposed DGS methods.} \label{tab:cap}
\begin{tabular}{c| c| c| c}
  \hline
    Dataset & Optical flow \cite{horn1981determining} & TSS method \cite{muhammad2022self}& DGS
  \\
  \hline
     CASIA-FASD  & 26 & 19 &  09 \\
      REPLAY-ATTACK & 18 & 13 &  05\\  
  \hline
\end{tabular}%
\end{table}

\begin{table}[]
\centering
\caption{The performance of the intra-database evaluation in terms of HTER $(\%)$ and EER $(\%)$ against state-of-the-art methods.} \label{tab:cap2}
\begin{tabular}{c|c|c|c|c}\hline & \multicolumn{2}{c|}{REPLAY-ATTACK} & \multicolumn{1}{c|}{CASIA}  & \multicolumn{1}{c}{MSU} \\  \hline
{Methods} & {EER$(\%)$}  & {HTER $(\%)$} & EER$(\%)$ & EER$(\%)$ \\ \hline
LBP+LDA \cite{de2013can} & 18.25  & - & 21.01 & - \\ 
IQA \cite{galbally2014face} & -  & - & 32.46 & - \\ 
CDD  \cite{yang2013face}  & 9.75  & - & 11.85 & - \\ 
 IDA \cite{wen2015face}   & - & - & 12.97 & 8.58 \\ 
 Color LBP \cite{boulkenafet2015face}  & 37.9  & 21.0 & 35.4 & 44.8 \\ 
 Color SURF \cite{boulkenafet2016face}   & 26.9  & 19.1  & 24.3 & 29.7 \\  
  Patch-CNN \cite{atoum2017face}   & 0.72 & -  & 4.44 & - \\ 
  Hybrid CNN \cite{muhammad2019faceb}   & - & -  & 0.02 & 0.04 \\ 
  SLRNN \cite{muhammad2019facea}   & - & -  & 0.01 & 0.02 \\ 
  SPMT + SSD \cite{song2019discriminative}   & 0.04 & 0.06 & 0.04 & - \\ 
   GFA-CNN \cite{tu2020learning}   & 0.30  & -  & 8.3 & 7.5 \\ 
    S-CNN \cite{quan2021progressive}  & 0.28  & -  & 0.53 & 0.18 \\ 
 \hline
DGS(CNN-BiLSTM) & \textbf{0.01}  & \textbf{0.02} & \textbf{0.00} & \textbf{0.01} \\ \hline
            \end{tabular}
\end{table}

\begin{table}[t]
  \begin{center}
    \caption{Intra-database evaluation on the four official protocols of the OULU-NPU database. Comparative results are obtained from \cite{muhammad2022self}.}
    \label{tab:oulu}
     \scalebox{0.7}{
    \begin{tabular}{l|l|r|r|r}
    \hline
      \textbf{Protocol} & \textbf{Method} & \textbf{APCER(\%)} & \textbf{BPCER(\%)} & \textbf{ACER(\%)}\\
      \hline
      \multirow{2}{*}{1} & DeepPixBiS \cite{george2019deep} & 0.8 & 0.0 & 0.4\\ 
      & GRADIANT  \cite{boulkenafet2017competition} & 1.3 & 12.5 & 6.9\\
       & Auxiliary  \cite{liu2018learning} & 1.6 & 1.6 & 1.6\\ 
       & FaceDs \cite{jourabloo2018face} & 1.2 & 1.7 & 1.5\\  
       & STASN \cite{yang2019face} & 1.2 & 2,5 & 1.9\\
       & DSGTD \cite{wang2020deep} & 2.0 & 0.0 & 1.0\\
        & CDCN \cite{yu2020searching} & 0.4 & 0.0 & 0.2\\ 
        & S-CNN \cite{quan2021progressive} & 0.6 & 0.0 & 0.4\\ 
      & TSS (ResNet) \cite{muhammad2022self} & 0.6 & 10.3 & 5.5\\ 
       & DGS with CNN  & 1.2 & 8.4 & 4.8\\ 
        & TSS (ResNet-BiLSTM)  \cite{muhammad2022self} & 0.0 & 0.2 & 0.1\\ 
       & DGS with CNN-BiLSTM  & 0.2 & 0.0 & \textbf {0.1}\\ 
      \hline
      \multirow{2}{*}{2}  & DeepPixBiS \cite{george2019deep} & 11.4 & 0.6 & 6.0\\ 
      & GRADIANT \cite{boulkenafet2017competition} & 3.1 & 1.9 & 2.5\\
      & Auxiliary  \cite{liu2018learning} & 2.7 & 2.7 & 2.7\\  
          & FaceDs \cite{jourabloo2018face} & 4.2 & 4.4 & 4.3\\ 
      & STASN \cite{yang2019face} & 4.2 & 0.3 & 2.2\\  
      & DSGTD \cite{wang2020deep} & 2.5 & 1.3 & 1.9\\
       & CDCN \cite{yu2020searching} & 1.8 & 0.8 & 1.3\\
        & S-CNN \cite{quan2021progressive} & 1.7 & 0.6 & 1.2\\ 
         & TSS (ResNet) \cite{muhammad2022self} & 2.0 & 2.1 & 2.1\\ 
              & DGS with CNN  & 2.3 & 1.5 & 1.9\\ 
                & TSS (ResNet-BiLSTM) \cite{muhammad2022self} & 0.4 & 0.8 & 0.6\\ 
       & DGS with CNN-BiLSTM  & 0.2 & 0.4 & \textbf {0.3}\\ 
      \hline
      \multirow{2}{*}{3} & DeepPixBiS \cite{george2019deep} & 11.7$\pm$19.6 & 10.6$\pm$14.1 & 11.1$\pm$9.4\\  
      & GRADIANT \cite{boulkenafet2017competition} & 2.6$\pm$3.9 & 5.0$\pm$5.3 & 3.8$\pm$2.4\\  
     & Auxiliary \cite{liu2018learning}  & 2.7$\pm$1.3 & 3.1$\pm$1.7 & 2.9$\pm$1.5\\  
     & FaceDs \cite{jourabloo2018face} & 4.0$\pm$1.8 & 3.8$\pm$1.2 & 3.6$\pm$1.6\\ 
      & STASN \cite{yang2019face} & 4.7$\pm$3.9 & 0.9$\pm$1.2 & 2.8$\pm$1.6\\  
      & DSGTD \cite{wang2020deep} & 3.2$\pm$2.0 & 2.2$\pm$1.0 & 2.7$\pm$0.6\\ 
       & CDCN \cite{yu2020searching} & 1.7$\pm$1.5 & 2.0$\pm$1.2 & 1.8$\pm$0.7\\ 
       & S-CNN \cite{quan2021progressive} & 1.5$\pm$0.9 & 2.2$\pm$1.0 & 1.7$\pm$0.8\\ 
      & TSS (ResNet) \cite{muhammad2022self} & 7.2$\pm$8.3& 3.9$\pm$3.4 & 5.5$\pm$3.0\\
        & DGS with CNN & 8.7$\pm$1.6& 2.2$\pm$1.1 & 5.4$\pm$1.5\\
         & TSS (ResNet-BiLSTM) \cite{muhammad2022self} & 2.5$\pm$1.8 & 0.5$\pm$0.6 & 1.5$\pm$0.8\\
       & DGS with CNN-BiLSTM  & 2.1$\pm$1.5 & 0.4$\pm$0.8 & \textbf {1.3$\pm$1.0}\\
      \hline
      \multirow{2}{*}{4} & DeepPixBiS \cite{george2019deep} & 36.7$\pm$29.7 & 13.3$\pm$14.1 & 25.0$\pm$12.7\\  
      & GRADIANT \cite{boulkenafet2017competition} & 5.0$\pm$4.5 & 15.0$\pm$7.1 & 10.0$\pm$5.0\\  
       & Auxiliary \cite{liu2018learning}  & 9.3$\pm$5.6 & 10.4$\pm$6.0 & 9.5$\pm$6.0\\ 
      & FaceDs \cite{jourabloo2018face} & 1.2$\pm$6.3 & 6.1$\pm$5.1 & 5.6$\pm$5.7\\ 
      & STASN \cite{yang2019face} & 6.7$\pm$10.6 & 8.3$\pm$8.4 & 7.5$\pm$4.7\\ 
      & DSGTD \cite{wang2020deep} & 6.7$\pm$7.5 & 3.3$\pm$4.1 & 5.0$\pm$2.2\\ 
       & CDCN \cite{yu2020searching} & 4.2$\pm$3.4 & 5.8$\pm$4.9 & 5.0$\pm$2.9\\ 
       & S-CNN \cite{quan2021progressive} & 5.2$\pm$2.0 & 4.6$\pm$4.1 & \textbf{4.8$\pm$2.0}\\ 
       & TSS (ResNet) \cite{muhammad2022self} & 5.7$\pm$4.5 & 16$\pm$13 & 10.8$\pm$5.3\\ 
              & DGS with CNN  & 6.4$\pm$5.9 & 14$\pm$11 & 10.2$\pm$5.0\\ 
              & TSS (ResNet-BiLSTM) \cite{muhammad2022self} & 4.7$\pm$10.5 & 9.2$\pm$10.4 & 7.1$\pm$5.3\\ 
       & DGS with CNN-BiLSTM  & 5.1$\pm$12.3 & 8.3$\pm$10.9 & 6.7$\pm$5.8\\ 
      \hline
    \end{tabular}}
 \end{center}
\end{table}

\begin{table}[t]
\begin{center}
\caption{Cross-database performance in terms of HTER $(\%)$ on the Replay-Attack and CASIA-FASD databases. Comparative results are obtained from \cite{muhammad2022self}.} \label{tab:xdb}
\begin{tabular}{c |c| c}
  \hline
    \multirow{2}{*}{Method} & Train CASIA-FASD  & Train Replay-Attack \\
      & Test Replay-Attack  & Test CASIA-FASD \\
  \hline
                   LBP \cite{maatta2011face}  & 47.0 &  39.6\\
                LBP-TOP \cite{de2014face}  & 49.7 &  60.6 \\
                  Color-LBP \cite{boulkenafet2015face}  & 30.3 &  37.7 \\
                  Motion-Mag \cite{bharadwaj2013computationally}  & 50.1 &  47.0 \\
                  Spectral cubes \cite{pinto2015face}   & 34.4 &  50.0\\
    Auxiliary \cite{liu2018learning}  & 27.6  &  28.4 \\
                   FaceDs \cite{jourabloo2018face}  & 28.5 & 41.1 \\
                   STASN \cite{yang2019face}   & 31.5 &  30.9 \\
                   DSGTD \cite{wang2020deep} & 17.0 & 22.8 \\
                      CDCN \cite{yu2020searching}  & 6.5 &  29.8\\
    TSS with ResNet \cite{muhammad2022self} & 30.4 & 39.9 \\
    DGS with ResNet (Ours) & 25.1 & 35.3 \\
    Self-supervised learning w/o BiLSTM & 22.1 & 36.8 \\
    Self-supervised learning \cite{muhammad2022self} & 5.9 & 15.2 \\
    Self-supervised learning (Ours) & \textbf{5.2} & \textbf{12.9} \\
  \hline
 \end{tabular}
\end{center}
\end{table}

\section{Experiments}
We conduct experiments on four major databases: Idiap Replay-Attack \cite{chingovska2012effectiveness}, CASIA Face Anti-Spoofing \cite{zhang2012face}, MSU Mobile Face Spoofing \cite{wen2015face}, and  OULU-NPU \cite{boulkenafet2017oulu}.  The Idiap Replay-Attack database contains $1,300$ videos, recorded with a built-in webcam of a Macbook Air laptop. The $1,300$ live and spoofed videos were then splitted in the following way: $60$ live samples and $300$ attacks under different lighting conditions as training set; $60$ live and $300$ attacks as development set; and $80$ real-accesses and $400$ attacks under various lighting conditions were used as a Test set. The enrollment set consists of $100$ real-accesses under different lighting conditions. The CASIA-FASD database consists of warped photo attacks, video-replay attacks, and cut-photo attacks. Overall, each subject comprised $12$ videos ($3$ genuine and $9$ fake), and the final database provides $600$ video clips. The MSU Mobile Face Spoof database contains $280$ videos and recorded from $35$ candidates using two kinds of cameras (laptop and Android phone). The video length of each video is at least nine seconds. For the Android smartphone camera, the videos were recorded with a resolution of $720 \times 480$, while for the laptop (MacBook Air $13$-inch), the videos were captured with a resolution of $640 \times 480$. Different kinds of presentation attack instruments (PAI), including prints, high definition laptop screens, and smartphone displays, were used to generate the attack presentations. One of the most challenging datasets is the OULU-NPU dataset that uses four protocols and consists of $55$ subjects ($15$ female and $40$ male), and $4950$ real access and attack videos. Protocol $1$ and Protocol $2$ are developed under different environmental conditions, namely illumination and background scene (e.g., unseen printers or displays). Protocol $3$ exploits a Leave One Camera Out (LOCO) protocol, to investigate the camera variation. Protocol $4$ combines all three protocols to simulate real-world operational conditions. For evaluation metrics, we report Half Total Error Rate (HTER), and Equal Error Rate (EER), by following the recently standardized ISO/IEC 30107-3 metrics on biometric presentation attack detection \cite{standard2017information}. For the Oulu dataset, the Average Classification Error Rate (ACER) is reported, which represents the mean of Bona Fide Presentation Classification Error Rate (BPCER) and Attack Presentation Classification Error Rate (APCER).

\subsection{Implementation details}
For assessing the performance of the proposed method, the experiments are conducted based on two approaches. In the first approach, the training and evaluation (testing) samples are used from the same dataset and referred to as an intra-database approach. In the second approach, a more realistic evaluation mechanism is used where the model is trained and tested on a completely unseen dataset called the cross-database, or inter-database approach. One of our aims in this paper is to evaluate the proposed DGS with both supervised and self-supervised paradigms. Therefore, in order to evaluate the performance of intra-database approach, the ResNet-101 \cite{he2016deep} is fine-tuned in stage 1 as shown in Fig.2, with stochastic gradient descent (SGD), mini-batch size $32$, validation frequency $30$, shuffle every-epoch, and learning rate of $0.0001$. An early stopping function \cite{prechelt1998early} is utilized to reduce the risk of over-fitting without setting the fixed epochs. All the input images (DGS) are resized to $224 \times 224$ and provided as input to the ResNet model without face cropping or data augmentation. The features are extracted in stage 1 from the last pooling layer of the fine-tuned model with a size of $2048$. We then use Bidirectional LSTM (BiLSTM) \cite{schuster1997bidirectional} in stage 2 that maps representations to space where the cross-entropy loss is applied. For BiLSTM, the Adam optimizer is utilized by fixing a learning rate of $0.0001$ with $500$ hidden layer dimension for all the intra-database protocols.

To show the generalization ability of the proposed approach in cross-database and self-supervised experiments, we keep the same parameters as mentioned above except the learning rate which was decreased to $0.001$. For self-supervised training, we first fine-tune the ResNet on unlabeled data, i.e., a set of different combinations (e.g.,$40, 50, 60$) for pretext task learning. Then, the model is fine-tuned on the supervised labels in the downstream task by replacing the new fully connected layer with the output size of $2$. Finally, BiLSTM takes the input of the last average pooling layer of the fine-tuned ResNet and performs the final detection. Initializing the network with the right weights remains always challenging because standard gradient descent from random initialization can hamper the learning of a BiLSTM network. Therefore, we use {\textit{He}} initializer \cite{he2015delving} for initializing recurrent weights that preform the best for all our experiments.

\subsection{Comparisons of execution times}
To evaluate the computational merit of the proposed DGS approach, we analyze the execution times of the global motion estimation \cite{muhammad2022self}, and optical flow \cite{horn1981determining} with our approach. In particular, given the video segments, we measure the frames per second (fps) for the above-mentioned methods of generating encoded segments for the temporal stream CNN. The results are reported in Table I. All these methods are implemented in a MATLAB environment by using a workstation with 3.5 GHz Intel Core i7-5930k and 64 GB RAM. The results demonstrate that DGS provides more than two times faster than the method based on global motion (TSS) and more than three times faster than the optical flow algorithm on the CASIA dataset. Similarly, DGS is three and two times faster for the REPLAY-ATTACK dataset respectively. 

\begin{figure}
\centerline{\includegraphics[width=08.90cm]{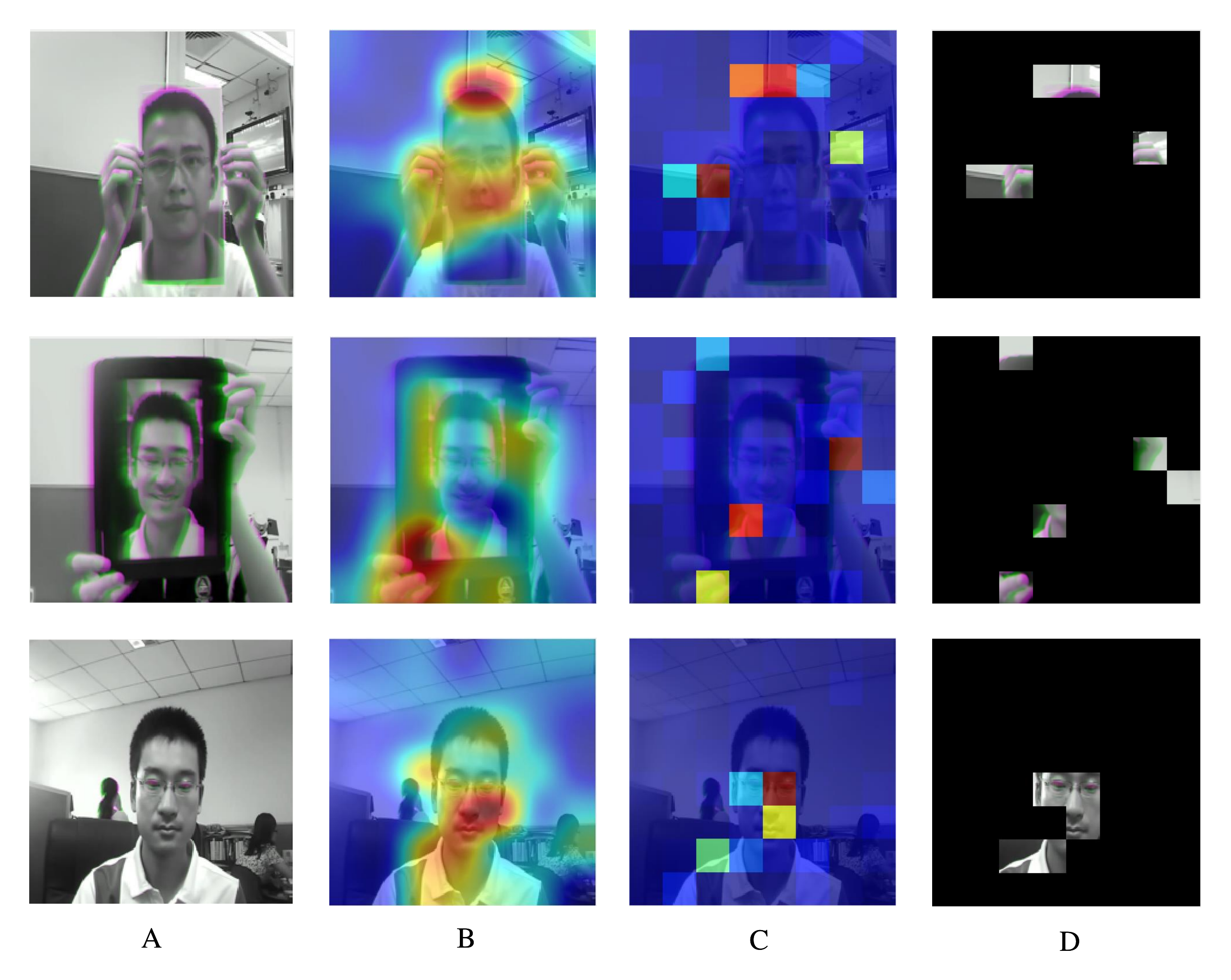}}
\caption{Image explanation using the Grad-CAM and LIME for DGS encoded videos corresponding to a print attack (first row), video-replay attack (second row) and real face (third row).}
\end{figure}
\subsection{Comparison against the state-of-the-art methods}
We investigate the effect of the DGS method in the intra-database scenario using all four datasets. Table II summarizes the results for the CASIA, REPLAY, and MSU datasets. The proposed DGS method in supervised settings provides state-of-the-art performance in comparison to the previous contemporary methods such as S-CNN \cite{quan2021progressive}. Specifically, our method achieves $0.01$, $0.00$, and $0.01$ EER for the Replay-Attack, CASIA, and MSU datasets, respectively. This is a remarkable improvement for the intra-database scenario. Table III illustrates the results of the OULU-NPU database by following the official evaluation protocols. We also report the performance without using BiLSTM. From these results, one can see that the proposed DGS method with the CNN-BiLSTM framework ranks first on protocols 1, 2, and 3 of the  OULU-NPU database. To verify the generalization ability to unseen spoofing attacks, we conduct experiments in the inter-database scenario and report the results in Table IV. Our DGS method performs significantly better than the work reported in \cite{muhammad2022self}, especially for the Replay-Attack dataset. One can observe that the proposed self-supervised learning improves more than three percent performance on the Replay-Attack dataset while slight drops in the performance for the CASIA dataset. However, BiLSTM further improves the performance when the proposed self-supervised learning stage is included in training the BiLSTM model. Thus, the proposed self-supervised learning helps to decrease the gap between unsupervised and supervised feature learning.

\subsection{Visualization and Analysis}
To observe the improvements brought by the DGS mechanism, we generate the class activation maps by using the Grad-CAM \cite{selvaraju2017grad} to see what exactly our model learns about real and spoofed faces. Example images from the real, video, and print attack categories are taken and visualized in the first column of Fig.4. The Grad-CAM heat-map emphasizes the paper artifacts, tablet edges, the mouth movements, and can be observed in the second column of Fig.4. Overall, we have a much more precise region of emphasis that discriminates real and attack classes. We hypothesize that DGS helps the model to classify the images correctly due to its intrinsic features, not a general region in the image. Similarly, in the third column of Fig.4., we illustrate the interpretability of the model using LIME (local interpretable model-agnostic explanations) \cite{ribeiro2016should} to understand the importance of the proposed DGS. One can see that a model keeps focusing on paper artifacts where the hand movement cues are valuable for the detection of the network. The tablet's edges provide salient information for the replay attack while the head movement or eye blinking contribute positively to separate live and spoofed faces. Finally, the last column of Fig.4 displays the masked images where only the most important superpixels are visible. Thus, based on the local explanations, we have determined the importance of DGS for PAD detection.

\section{Conclusions}
In this paper, we addressed the face PAD issue by proposing a self-supervised learning method that leverages dynamic grayscale snippets (DGS) and use different temporal lengths to self-generate a supervisory signal for the downstream task. Since the motion-based cues are naturally available in the video sequences, we investigated how well the brightness discrepancy of the corresponding pixels in DGS can describe the intrinsic disparities between live and spoofed faces. Furthermore, the temporal redundancy in PAD videos is eliminated and the proposed motion representation (DGS) can be adopted for the 2D temporal stream CNNs. Extensive experiments on four datasets demonstrate that our proposed method is robust in both intra-database and inter-database testing scenarios. Our future work includes utilizing video classification methods to represent videos in a more discriminative form and formulate self-supervised learning strategies to further improve the robustness of PAD methods.

\bibliographystyle{ieeetr}

\bibliography{references}  






\end{document}